\documentclass[letterpaper, 10 pt, conference]{ieeeconf}  

\IEEEoverridecommandlockouts                              

\usepackage{graphics} 
\usepackage{epsfig} 
\usepackage{mathptmx} 
\usepackage{times} 
\usepackage{amsmath} 
\usepackage{amssymb}  
 \usepackage[linesnumbered,ruled]{algorithm2e}
\usepackage[]{algorithm2e}
\usepackage{cleveref}
\usepackage{siunitx}
\usepackage{ntheorem}
\usepackage{tabularx}
\newtheorem{hyp}{Hypothesis}

\crefname{figure}{Fig.}{Fig.}
\Crefname{figure}{Fig.}{Fig.}

\usepackage{soul}
\usepackage{color}

\title{\LARGE \bf
Experimental Evaluation of Human Motion Prediction \\ Toward Safe and Efficient Human Robot Collaboration
}

\author{Weiye Zhao, Liting Sun, Changliu Liu and Masayoshi Tomizuka
\thanks{This work was supported by National Science Foundation (Award \#1734109)}\thanks{W. Zhao and C. Liu are with the Robotic Institute, Carnegie Mellon University, Pittsburgh, PA 15213 USA (e-mail: \tt\small weiyezha, cliu6@andrew.cmu.edu).}
\thanks{L. Sun and M. Tomizuka are with the Department of Mechanical Engineering, University of California, Berkeley, CA 94720 USA (e-mail: \tt\small liting, tomizuka@berkeley.edu).
}}

\begin{document}

\maketitle
\thispagestyle{empty}
\pagestyle{empty}

\begin{abstract}
Human motion prediction is non-trivial in modern industrial settings. Accurate prediction of human motion can not only improve efficiency in human robot collaboration, but also enhance human safety in close proximity to robots. Among existing prediction models, the parameterization and identification methods of those models vary. It remains unclear what is the necessary parameterization of a prediction model, whether online adaptation of the model is necessary, and whether prediction can help improve safety and efficiency during human robot collaboration. These problems result from the difficulty to quantitatively evaluate various prediction models in a closed-loop fashion in real human-robot interaction settings.
This paper develops a method to evaluate the closed-loop performance of different prediction models. In particular, we compare models with different parameterizations and models with or without online parameter adaptation.  Extensive experiments were conducted on a human robot collaboration platform. The experimental results demonstrated that human motion prediction significantly enhanced the collaboration efficiency and human safety. Adaptable prediction models that were parameterized by neural networks achieved the best performance. 
\end{abstract}

\section{INTRODUCTION}
With the revolutionary development of manufacturing automation, there are increasing manufacturing tasks that are tedious and inefficient for human to perform~\cite{mainprice2013human}. Effective human-robot collaboration (HRC) has drawn increasing interest across large varieties of fields~\cite{yamazaki2012home}~\cite{knight2013smart}~\cite{diftler2011robonaut}. Moreover, many challenging HRC tasks, such as aircraft and electronics assembly~\cite{lasota2015analyzing}, require human workers to work in close proximity to robots. To enable safe and efficient HRC, it is crucial for robots to be aware of current and future human movements, so that robots can quickly adapt their behavior to better collaborate with human workers~\cite{cheng2018human}. However, it is inherently difficult to predict human motion, since human motion is stochastic in nature and subjected to nonlinear dynamics~\cite{peng2014assessing}. Human behavior is time-varying and individual differences are prominent. A prediction model that works for one person may not be applicable to another, or even the same person at a different time.

A prediction model can be parameterized in multiple ways, i.e., linear regression model, supported vector machine, Gassian mixture model, hidden Markov model~\cite{wu2014leveraging}, feed-forward neural network, recurrent neural networks (RNNs)~\cite{ghosh2017learning}, and etc. The parameters of a prediction model can either be fixed or online adaptable.  

In terms of performance evaluation of a prediction model, the majority of existing works focus on prediction accuracy. Few of them evaluate the effectiveness of the prediction in real world experiments with human-robot systems. 
Without fully evaluating the effects of human motion prediction through real world experiments involving human, it is hard to determine whether a prediction model  would lead to safe and efficient human robot collaboration. 

This paper introduces a series of human-in-the-loop co-robot experiments to compare different prediction models. The co-robot platform is based on the safe and efficient robot collaborative system (SERoCS)~\cite{liu2018serocs}. We investigate the following problems:
\begin{enumerate}
\item whether a complex parameterization (e.g., using a neural network) of a prediction model is necessary;
\item whether online adaptation of a prediction model is necessary;
\item whether active prediction improves safety and efficiency of human-robot collaboration.
\end{enumerate}
To answer the questions, we compare the following four types of prediction models: 
\begin{enumerate}
    \item a linear regression model without adaptation,
    \item a linear regression model with adaptation,
    \item a neural network model without adaptation,
    \item a neural network model with adaptation, called a semi-adaptable neural network model.
\end{enumerate}

In the experiments, the baseline for comparison is chosen to be the performance without active prediction, in which robot only considers real-time human constraints. The metrics for performance evaluation are defined by quantifying safety and efficiency during human robot collaboration. The experimental results demonstrate that safety score is doubled with active prediction. Prediction models with complex parameterizations and online adaptations achieve higher efficiency scores compared to prediction models without those. Among the four methods, the semi-adaptable neural network model has the best performance.  

The remainder of the paper is organized as follows. Section II formulates the human motion prediction problem and describes the four motion prediction methods mathematically. Section III proposes three hypotheses about the effects of motion prediction on human robot collaboration. Section IV describes the details of experimental setup. Section V shows the performance of the four methods. Section VI discusses how the experimental results verify the three hypotheses and Section VII concludes the paper.

\section{Human Motion Prediction Methods}
Following the formulation in~\cite{cheng2018human}, the transition model of human motion on a selected joint can be defined as: 
\begin{equation}
{\mathbf{x}(k+1)}  = f(\mathbf{x}^*(k), a) + w_k,
\label{eq: dynamic model}
\end{equation}
where $\mathbf{x}(k+1)\in \mathbb{R}^{3M}$ denotes human's $M$-step future trajectory of the joint, in which three dimensional joint position in Cartesian-space is included for each future step and $M$ is the prediction horizon. $\mathbf{x}^*(k) \in \mathbb{R}^{3N}$ denotes human's past $N$-step trajectory of the joint. $a \in \mathbb{N}^1$ is a discrete action label representing different motion categories. $w_k \in \mathbb{R}^{3M}$ is a zero-mean white Gaussian noise. The function $f(\mathbf{x}^*(k), a): \mathbb{R}^{3N} \times \mathbb{N}^1 \to \mathbb{R}^{3M}$ encodes the transition of the human motion, which takes historical joint trajectory and current action label as inputs, and outputs the future positions of the joints. 

The following subsections summarize the four human motion prediction methods compared in the study. The four methods can be divided into two categories: offline schemes and online schemes. 1) Offline schemes: linear regression model without adaptation and neural network model without adaptation. 2) Online schemes: linear regression model with adaptation, and neural network model with adaptation. 

\subsection{Linear Regression Model without Adaption}
A human is a complex dynamic system~\cite{martin1999control}, whose dynamics \eqref{eq: dynamic model} can be approximated by a linear regression model~\cite{hasler2009statistical}: 
\begin{equation}
	\mathbf{x}(k+1) =\Phi_k \theta+w.\label{eq: identification model}
\end{equation}

Here $\mathbf{x}(k+1)$ is the output of the system, and $\Phi_k$ is the regressor vector that contains the past observations of the system at time step $k$, and $\theta$ is the unknown model parameters. $w$ is a white Gaussian stochastic noise. In terms of human motion prediction, $\mathbf{x}(k+1) \in\mathbb{R}^{3M}$ represents the future motion states, while the regressor vector $\Phi_k \in\mathbb{R}^{3N+1}$ is an augmented vector containing the past states $\mathbf{x}^*(k)$ and the action label $a$. 

The parameter $\theta$ needs to be identified from human motion transition dataset. Linear regression is usually solved using the least square method~\cite{sen1976line}, which is inefficient since it involves large scale matrix inverse. To speed up the computation, we adopt stochastic gradient descent (SGD) to perform the multi-variate linear regression, which has guaranteed convergence~\cite{liu2010convergence}. The main steps of the SGD algorithm can be summarized as that we first initialize parameter $\theta$ and white noise $v$. Next, for $k$-th motion trainsition sample $(\bar{\Phi}_k,\bar{\mathbf{x}}(k+1))$ in dataset, we construct the error function as:
\begin{equation}
E_k(\theta) = \|\bar{\mathbf{x}}(k+1) - \bar{\Phi}_k \theta\|^2
\label{eq: error function}
\end{equation}
Then the parameter is updated using the following rule:
\begin{equation}
\theta^{(i+1)} = \theta^{(i)} - \eta\frac{1}{m}\sum_{k=1}^{m}\nabla E_k(\theta^{(i)})
\label{eq: descent}
\end{equation}
where $\theta^{(i)}$ is the parameter in $i$-th iteration. $\theta^{(i)}$ converges when $i$ approaches infinity.

\subsection{Neural Network Model without Adaptation}
Since human's motion is inherently highly nonlinear, complex parameterization is favorable for motion prediction models. RNNs receives increasing interests in human motion prediction recently. However, RNNs is hard to train, many attempts to accelerate the speed have failed~\cite{cheng2018human}. Moreover, since RNNs models are fixed, their recurrent network architectures are not compatible with online adaptation schemes. For comparison purpose, we need a feed-forward model with complex parameterization that is online adaptable. Hence, we use Artificial Neural Networks (ANNs) to approximate the motion transition model $f$ in \eqref{eq: dynamic model}, which is fast to train and widely used in modeling human motion~\cite{bataineh2016neural}.

To train the transition model $f$, we define an $n$-layer ANN with ReLU activation function which takes the positive part of the input to a neuron as:
\begin{align}
f(\mathbf{x}^*(k), a)  = W^T \max(0, g(U, s_k)) + \epsilon(s_k),\label{eq: T1 NN}
\end{align}
where $s_k = [\mathbf{x}^*(k)^T, a,1]^T \in \mathbb{R}^{3N+2}$ is the input vector, in which $1$ denotes a bias term. $g$ denotes $(n-1)$ - layer neural network, whose weights are packed in $U$. $\epsilon(s_k) \in \mathbb{R}^{3M}$ is the function reconstruction error, which goes to zero when the neural network is fully trained. $W\in\mathbb{R}^{n_h\times 3M}$ is the last layer parameter weights, where $n_h \in \mathbb{N} $ is the number of neurons in the hidden layer of the neural network~\cite{ravichandar2017human}. We also deploy stochastic gradient descent to train ANN model following the same procedure as \eqref{eq: error function} and \eqref{eq: descent}.

\subsection{Linear Regression Model with Adaptation}
The prominent difficulty in human motion prediction lies in the individual difference. A fixed linear regression model as \eqref{eq: identification model} is far from satisfying in accommodating the stochasity  of human motion. To address the problem, We adapt the parameters of the linear regression model using the recursive least square parameter adaptation algorithm (RLS-PAA)~\cite{goodwin2014adaptive}. Rewrite the regression model \eqref{eq: identification model} as
\begin{equation}
	\mathbf{x}(k+1) =\Phi_k \theta_k+w_k,\label{eq:Transformed LTV}
\end{equation}
where $\theta_k$ and $w_k$ denote the parameter and white noise at time step $k$. Let $\hat{\theta}_k$ denotes the parameter estimate at time step $k$, and let $\tilde{\theta}_k=\theta_k-\hat{\theta}_k$ be the parameter estimation error. We define the \textit{a priori} estimate of the state and the estimation error as:
\begin{align}
	\hat{\mathbf{x}}\left(k+1|k\right) =&\Phi_k\hat{\theta}_k,\label{eq:prediction}\\
	\tilde{\mathbf{x}}\left(k+1|k\right) =&\Phi_k\tilde{\theta}_k+w_k.
\end{align}

The main steps of RLS-PAA are to iteratively update the parameter estimation $\hat{\theta}_k$ and predict  $\mathbf{x}(k+1)$ when new measurements become available. The parameter update rule of RLS-PAA can be summarized as~\cite{goodwin2014adaptive}:
\begin{equation}
	\hat{\theta}_{k+1}=\hat{\theta}_k+F_k\Phi^{T}_k\tilde{\mathbf{x}}\left(k+1|k\right),
\label{eq:update_theta}
\end{equation}
where $F_k$ is the learning gain updated by:
\begin{equation}
F_{k+1}= \frac{1}{\lambda_1(k)}[F_k - \lambda_2(k)\frac{F_k\Phi_k\Phi^T_kF_k}{\lambda_1(k)+\lambda_2(k)\Phi^T_kF_k\Phi_k}]\label{eq:update_F},
\end{equation}
where $0<\lambda_1(k)\leq1$ and $0<\lambda_2(k)\leq2$.
Typical choices for $\lambda_1(k)$ and $\lambda_2(k)$ are:
\begin{enumerate}
  \item $\lambda_1(k) = 1$ and $\lambda_2(k) = 1$ for standard typical least squares gain.
  \item $0 < \lambda_1(k) < 1$ and $\lambda_2(k) = 1$ for least squares gain with forgetting factor.
   \item $\lambda_1(k) = 1$ and $\lambda_2(k) = 0$ for constant adaptation gain.
\end{enumerate}

\subsection{Neural Network Model with Adaptation}
The last-layer weight $W$ in the feed-forward neural network \eqref{eq: T1 NN} is also adaptable online using RLS-PAA~\cite{cheng2018human}. We call the new model the semi-adaptable neural network.
The semi-adaptable neural network is more computational effective than RNNs and can adapt time-varying human motion online. This method requires that the neural network be pre-trained offline. The pre-trained network will serve as an effective feature extractor when the last layer is removed~\cite{athiwaratkun2015feature}. To accommodate time-varying behaviors and individual differences in human motion, we just need to adjust $W$, weights in the last layer of the neural network. 

To apply RLS-PAA on the adaptation of $W$ in \eqref{eq: T1 NN}, we can stack all the column vectors of the matrix $W$ to form a new parameter vector  $\theta\in\mathbb{R}^{3Mn_h}$. $\theta_k$ denotes the value of the parameter vector at time step $k$. To represent the extracted features, we define a new data matrix $\Phi_k \in\mathbb{R}^{3M\times 3Mn_h}$ as a diagonal concatenation of $M$ pieces of $\max(0, g(U,s_k))$. Using $\Phi_k$ and $\theta_k$, then \eqref{eq: dynamic model} and \eqref{eq: T1 NN} can be written into the same form as \eqref{eq:Transformed LTV}. The 
procedure of semi-adaptable neural network is summarized in \Cref{ANN}.
\begin{algorithm}
    \SetKwInOut{Input}{Input}
    \SetKwInOut{Output}{Output}
    \SetKwInOut{Variables}{Variables}
    \SetKwInOut{Initialization}{Initialization}
    \Input{Offline trained neural network \eqref{eq: T1 NN} with $g$, $U$ and $W$}
    \Output{future trajectory $\mathbf{x}(k+1)$}
    \Variables{Adaptation gain $F$, neural network last layer parameters $\theta$, variance of zero-mean white Gaussian noise $Var(w_k)$}
    \Initialization{$F=1000\mathcal{I}$,
    $\theta=$ column stack of $W$, $\lambda_1=0.998$, $\lambda_2=1$}
    \While{True}{
    Wait for a new joint position $p$ captured by Kinect and current action label $a$ from action recognition module\;
    Construct $s_k=[\mathbf{x}^*(k), a,1]^T$\;
    Obtain $\Phi(k)$ by diagonal concatenation of $max(0, g(U,s_k))$\;
    Update $F$ by (\ref{eq:update_F})\;
    Adapt the parameters $\theta$ in last layer of neural network by (\ref{eq:update_theta})\;
    Calculate future joint trajectory $x(k+1)$ by (\ref{eq:Transformed LTV})\;
    send $x(k+1)$ to robot control. 
    }
    \caption{Semi-adaptable neural network for human motion prediction}
    \label{ANN}
\end{algorithm}

\section{Hypothesis}
Based on the prediction models discussed in the previous section, we anticipate that active prediction will affect both safety and efficiency during human robot collaboration, and the effects varies for models with different parameterizations and models with or without online adaptation. Here we propose three main hypotheses, which will be verified in the experiments to be discussed in the following sections:

\begin{hyp}[Prediction Accuracy]
Online adaptation of a prediction model improves the prediction accuracy. The prediction accuracy is higher for models that can encode nonlinear features. 
\end{hyp}

\begin{hyp}[Prediction and Safety]
Active human motion prediction  enables the robot motion planner to take human tendency into consideration, which improves human safety.  
\end{hyp}

\begin{hyp}[Prediction and Efficiency]
Collaboration efficiency should be higher for prediction models with higher prediction accuracy (e.g., adaptable prediction models).
\end{hyp}

\section{Experiment Design}
To quantitatively evaluate the effects of motion prediction on human robot collaboration, we conduct a series of experiments in which a human works in close proximity to a robot while the robot is performing predefined tasks.

\subsection{Experiment Setup}
\begin{figure}
\begin{center}
\includegraphics[scale=0.45]{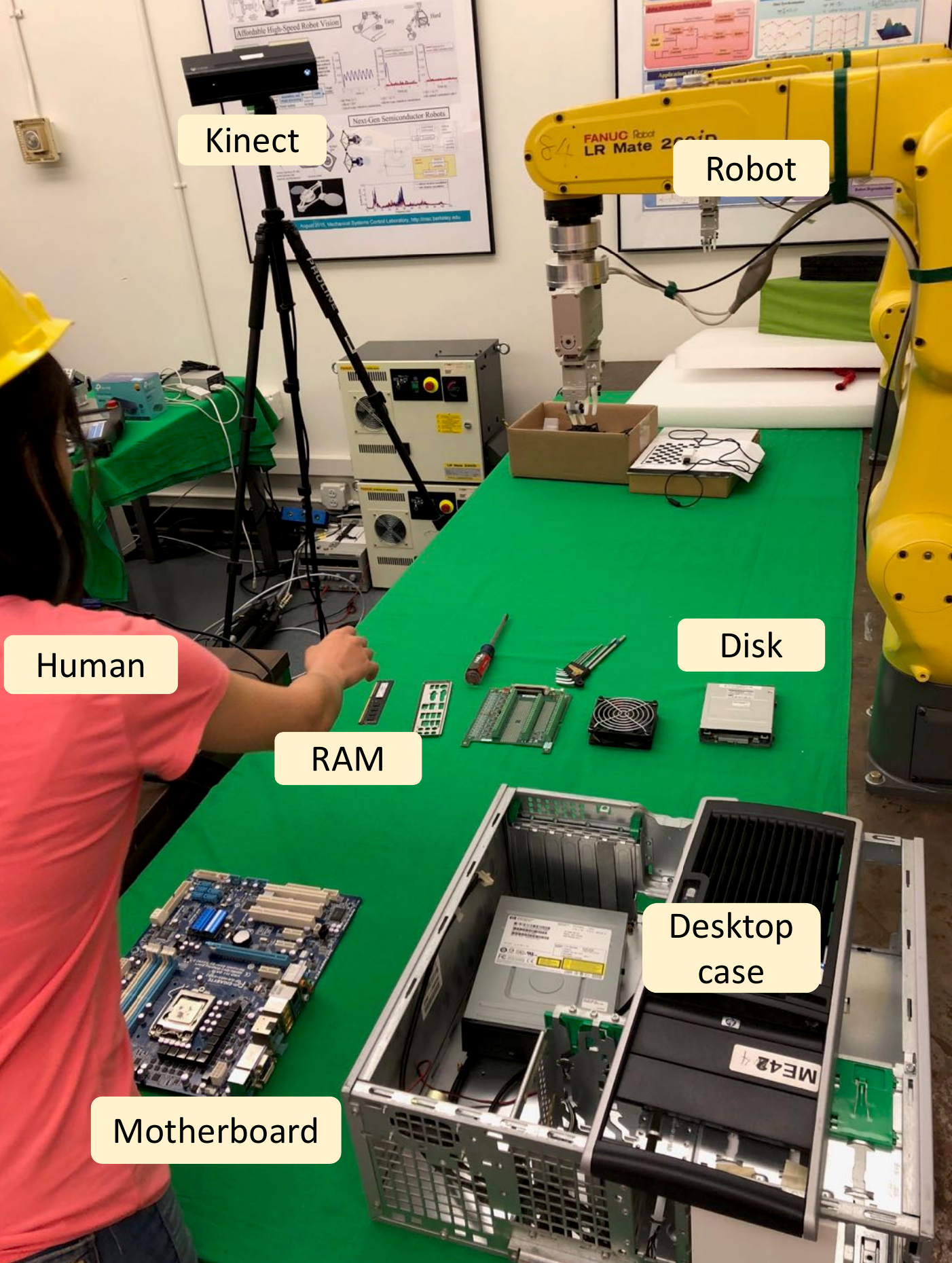}
\caption{Experimental setup for evaluation of different prediction models.}
\label{setup}
\end{center}
\end{figure}

The experiment platform is shown in \Cref{setup}. The robot manipulator is FANUC LR Mate 200iD/7L, a 6-degree-of-freedom industrial robot. There is one Kinect sensor to monitor the dynamic environment. We track the trajectory of the human's right wrist. The system update frequency is approximately \SI{20}{\hertz}. All the experiments are implemented in Matlab 2016 platform on a Windows desktop with \SI{2.7}{\giga\hertz} Intel Core i5 Processor and 16 GB RAM. We also deploy robot controller on the Simulink RealTime target. 

\subsection{Safe and Efficient Robot Collaborative Systems}
The experiment is built upon the safe and efficient robot collaboration system (SERoCS)~\cite{liu2018serocs}. SERoCS consists of three modules: (T1) the robust cognition module for environment monitoring and prediction, (T2) the optimal task planning module for efficient human-robot collaboration, and (T3) the motion planning and control module for safe human-robot interaction. In the experiment, the robot is only required to track a simple trajectory, which does not require task planning in T2. We then close the loop with only T1 and T3. The input of T1 is the real time visual information, while the output of T1 consists of the past human states $\mathbf{x}^*(k)$, and the predicted trajectory $\hat{\mathbf{x}}(k+1\mid k)$. In the comparison experiments, we use the four prediction models described in previous Section II as the prediction algorithm in T1.

We also deploy the T3 module to plan and control the robot motion. The module input is the environment information and the prediction of human motion from T1, while the output is the desired robot motion trajectory $\mathbf{x}_R$. The robot is equipped with 1) a short-term safety-oriented planner based on the safe set algorithm~\cite{liu2014control} and 2) a long-term efficiency-oriented planner based on the convex feasible set algorithm~\cite{liu2018convex}. The two planners run in parallel.

\begin{figure}
\begin{center}
\includegraphics[scale=0.45]{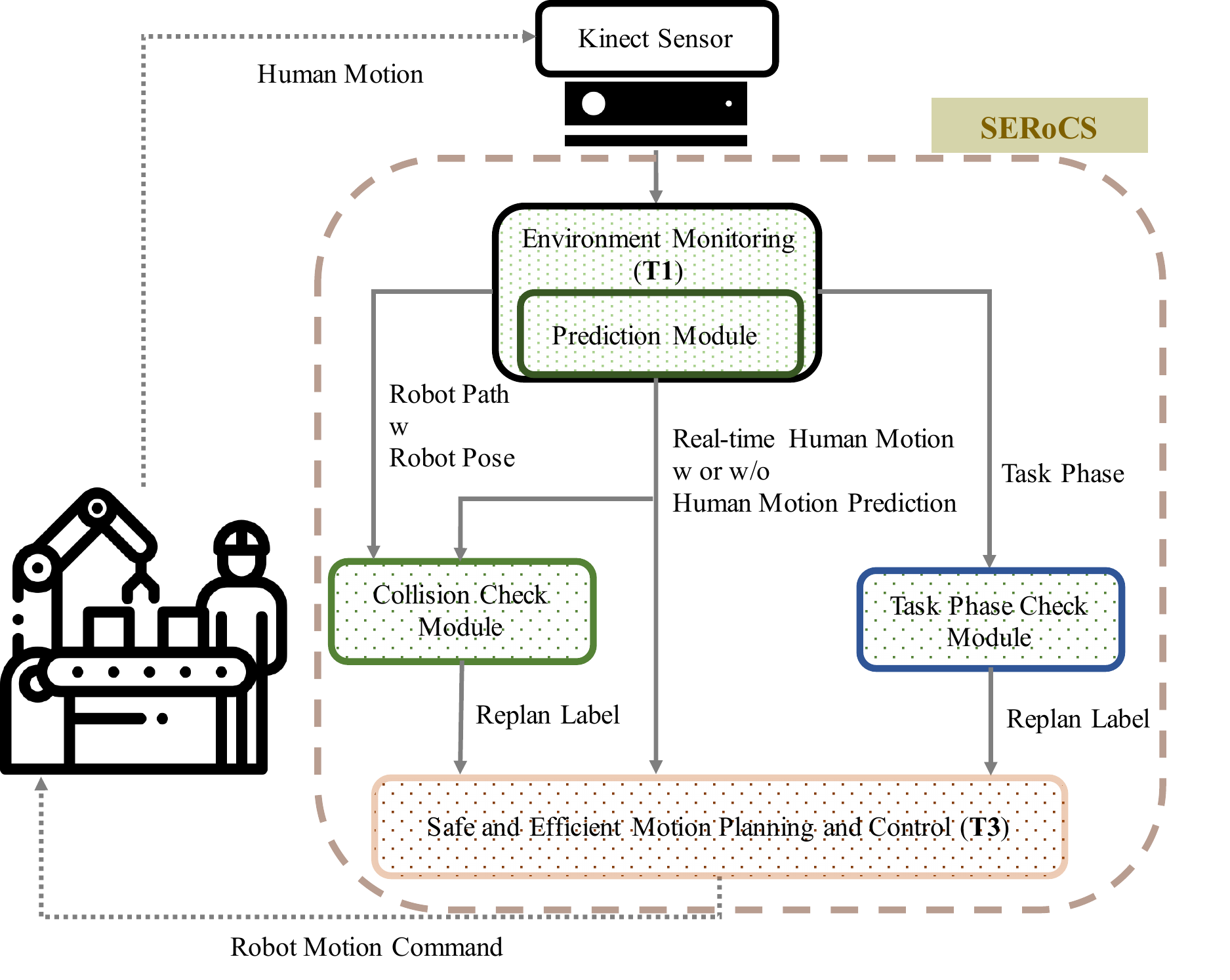}
\caption{Safe and efficient robot collaboration system (SERoCS).}
\label{serocs}
\end{center}
\end{figure}

The close-loop execution of the human robot collaboration system is summarized below, which is also shown in \cref{serocs}. 
\begin{enumerate}
	\item The T1 module estimates the  past human states $\mathbf{x}^*(k)$ and predicts the future trajectory $\hat{\mathbf{x}}(k+1\mid k)$ from the sensory data.
	\item The T3 module plans the future robot trajectory given information from the T1 module. The short term planner runs in receding horizon for \SI{1}{\hertz}. The long term planner only replans in two scenarios: when a new task is specified, or when the distance between the human and the robot is below a threshold. 
	\item The planned trajectory is send to the robot hardware for execution. New states of the system will be obtained in the next time step, and the steps 1) -3) repeat until experiments end.
\end{enumerate}

\subsection{Human-Robot Collaboration Tasks}
To effectively compare different types of motion prediction models, we need to design  human-robot interactive tasks considering the following features.

\subsubsection{Variety of human motions} 
To simulate the real factory scene, the improvement brought by motion prediction should be robust facing more than one type of complex human motions. Therefore, the task needs to emphasize the variety of human motions. 

\subsubsection{Responsiveness of the robot}
We want to test if the robot can operate safely and efficiently in a worst case scenario, i.e., when the human completely ignores the robot. In the way, the robot needs to react quickly to meet the human's needs.

\subsubsection{Repeatability}
All experiments should be repeatable. Robot should be required to perform the same task at each trial. The same motion planner should be applied. Different human motion types should also be designed and fixed across comparison experiments with different motion prediction models. 

\begin{figure}
\begin{center}
\includegraphics[scale=0.5]{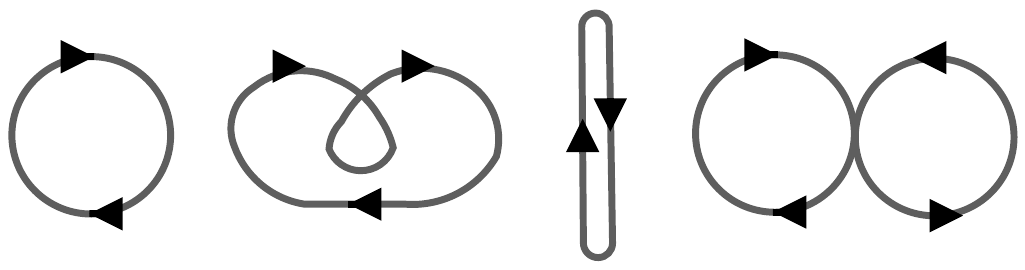}
\caption{Four predefined motions for human right hand.}
\label{motion}
\end{center}
\end{figure}

To meet these requirements, we design the tasks following a factory scenario and focus on the most common "Fetching" operation. As shown in \cref{setup}, we allocate a Disk and a RAM for the robot to fetch, starting from its idle pose. To emphasize the variety of human motions, we design four different motion patterns, as shown in \cref{motion}. Four motion patterns represent four different action labels $a$ in \cref{eq: dynamic model}. The action label is pre-defined across experiments. To test the responsiveness of the robot, we instruct the human in the experiments not to respond to the robot as long as safety is under control with emergency brakes. To guarantee the repeatability, we fix the locations of the Disk and RAM across different experiments. 

\subsection{Experiment Procedure}
\subsubsection{Prediction models}
We compare four different human motion prediction models: 1) Linear Regression Model without Adaptation, 2) Neural Network Model without Adaptation, 3) Linear Regression Model with Adaptation, and 4) Neural Network Model with Adaptation.

\subsubsection{Data collection}
In order to verify the effects of different motion prediction models, it is necessary to pre-train the prediction models before using them in online execution. We collected human motion data. For each type of motion, $30$ trajectories are obtained, and are used for four prediction models training. To smooth the noisy trajectories, we use a low-pass filter $p_s(k) = 0.6\hat{p}(k-1) + 0.4\hat{p}(k)$. $p_s(k)\in \mathbb{R}^3$ is the smoothed position of the joint at time step $k$, which is the weighted average of joint positions $\hat{p}(k-1)$ and $\hat{p}(k-2)$ measured at time step $k-1$ and $k-2$. We also set the number of past and future joint positions $N$ and $M$ both to be $3$. 

\subsubsection{Pre-computation}
We pre-train the prediction models using the collected data. For Linear Regression Model, we use a $10\times9$ matrix to represent transformation parameter $\theta$, and apply SGD to optimize $\theta$. For Neural Network Model, we apply a 3-layer neural network with $40$ nodes in the hidden layer.  The number of nodes in the input layer and the output layer is $11$ and $9$, respectively. The loss function is set to be L2 loss. All the learning rates are set to $0.001$ and the number of epochs is $100$.  Note that the parameters of RLS model and semi-adaptable neural network model can be initiated with parameters of the pre-trained linear regression model and offline neural network model, respectively. 

\subsubsection{Experimental validation}
After the pre-trained phase, we start human robot interactions experiments by substituting the prediction module with different prediction models. For each human motion, $20$ independent trials are conducted, and $80$ trials are conducted for each prediction model. We also conduct $80$ counterpart trials without prediction module as ground truth. Experiment data is also collected, including human position series, robot position series, human motion prediction series, and videos. 

\subsection{Evaluation Metrics}
\subsubsection{Prediction}
Accuracy is a good indicator for prediction performance. We define the average prediction error using average distance between the predicted trajectory and ground truth trajectory. A smaller average prediction error implies better prediction performance.
\subsubsection{Safety}
Robot is supposed to keep the proper distance from human to avoid potential collision. We define the average closet distance between the human and the robot as the safety index for each trial:
\begin{equation}
\text{Safety} = \frac{1}{n}\sum_{i=1}^{n}(\min(Dist(H_i,R_i))),
\label{eq: safety}
\end{equation}
where $n$ is the sample frame number for a trial, $H_i$ and $R_i$ denote human pose and robot pose at frame $i$, respectively. A higher safety index means the robot is farther from the human, hence safer. 
\subsubsection{Efficiency}
Proper prediction of human motion can make the robot escape in advance when human is approaching, and continue with its task when the human tends to get away. In other words, good motion prediction can improve the efficiency of robot motion by making the robot keep as close to its target as possible. Denote the average robot-target distance without human interference as ground truth $\mathbb{D}_{RT}$.  
We define the efficiency index by comparing the average distance between the robot and its target with the ground truth $\mathbb{D}_{RT}$: 
\begin{equation}
\text{Efficiency} = \frac{\mathbb{D}_{RT}}{\frac{1}{n}\sum_{i=1}^{n}(Dist(R_i, T))}
\label{eq: safety},
\end{equation}
where $T$ denotes the target position. A higher efficiency index indicates that the robot completes tasks more efficiently.

\section {Results}
\begin{figure}[b]
\begin{center}
\includegraphics[scale=0.45]{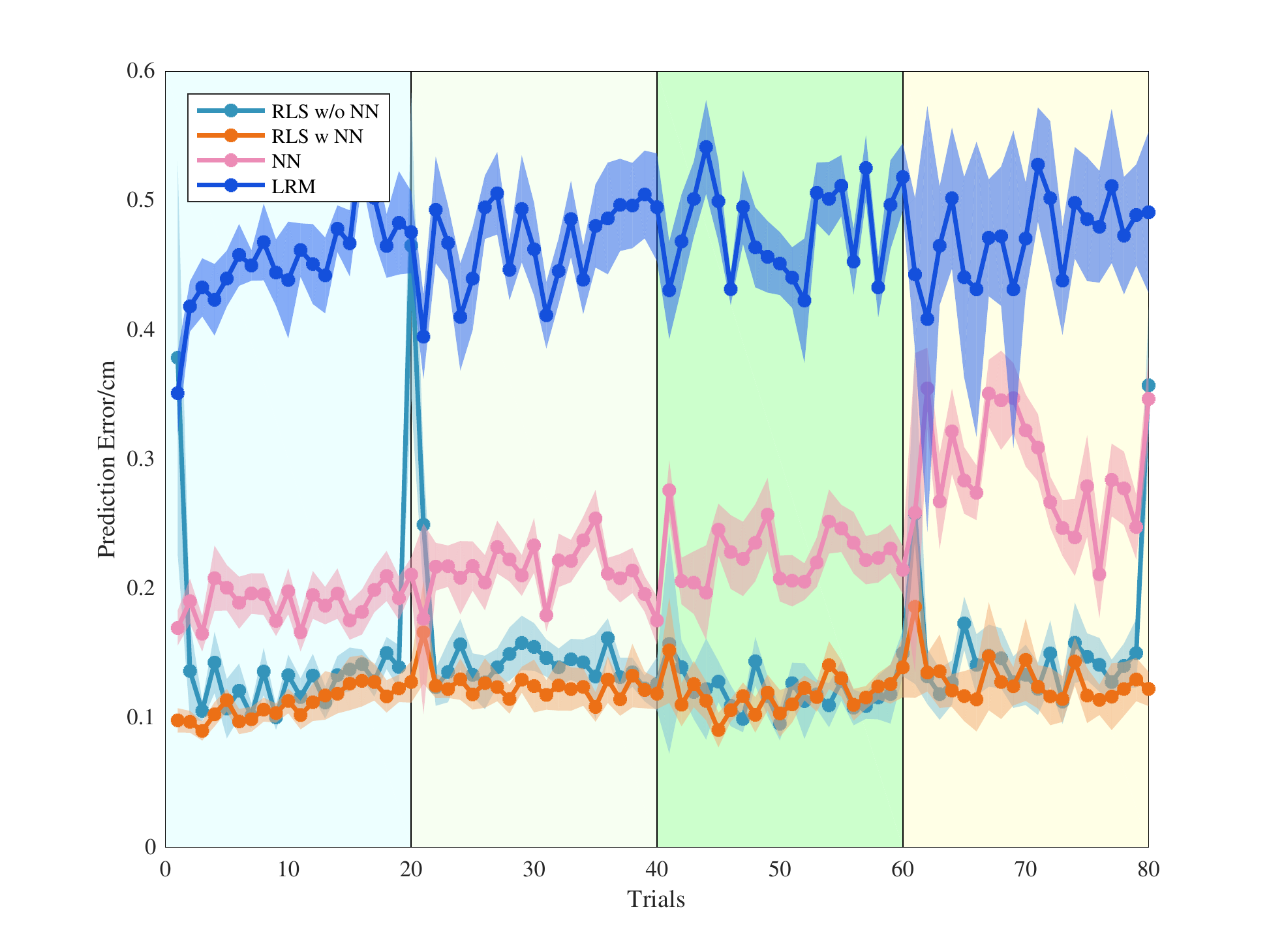}
\caption{Prediction error comparison among four different motion prediction algorithms on 80 trials. The bold lines are averaged over all sample points for each trial, the shaded area presents one standard deviation. rectangle backgrounds with different color denote the different motion classes. }
\label{err}
\end{center}
\end{figure}

\subsection{Prediction accuracy}
We first evaluate the performance of the four prediction methods according to the prediction error as shown in \Cref{err}. Semi-adaptable neural network results in much smaller prediction error and less wild standard deviation on majority of 80 trials. The statistics are also summarized in \Cref{summary_er}. 
\begin{table}
\caption{Mean and variance of the prediction error for four motion prediction models.}
 \centering
   \begin{tabular}{p{1.5cm}p{2.5cm}p{2.5cm}}
   \hline
   & prediction error mean \newline (\si{\meter}) & prediction error variance \newline (\si{\square\meter})\\
   \hline
   RLS w/o NN & 0.1447 & 0.0032\\

   RLS w NN & \bf{0.1209} & \bf{0.0002}\\

   NN  & 0.2304 &0.0021\\

   LTM & 0.4678& 0.0012 \\

   \hline
   \end{tabular}
 
 \label{summary_er}
\end{table}

\subsection{Safety and efficiency}
\begin{figure}
\begin{center}
\includegraphics[scale=0.45]{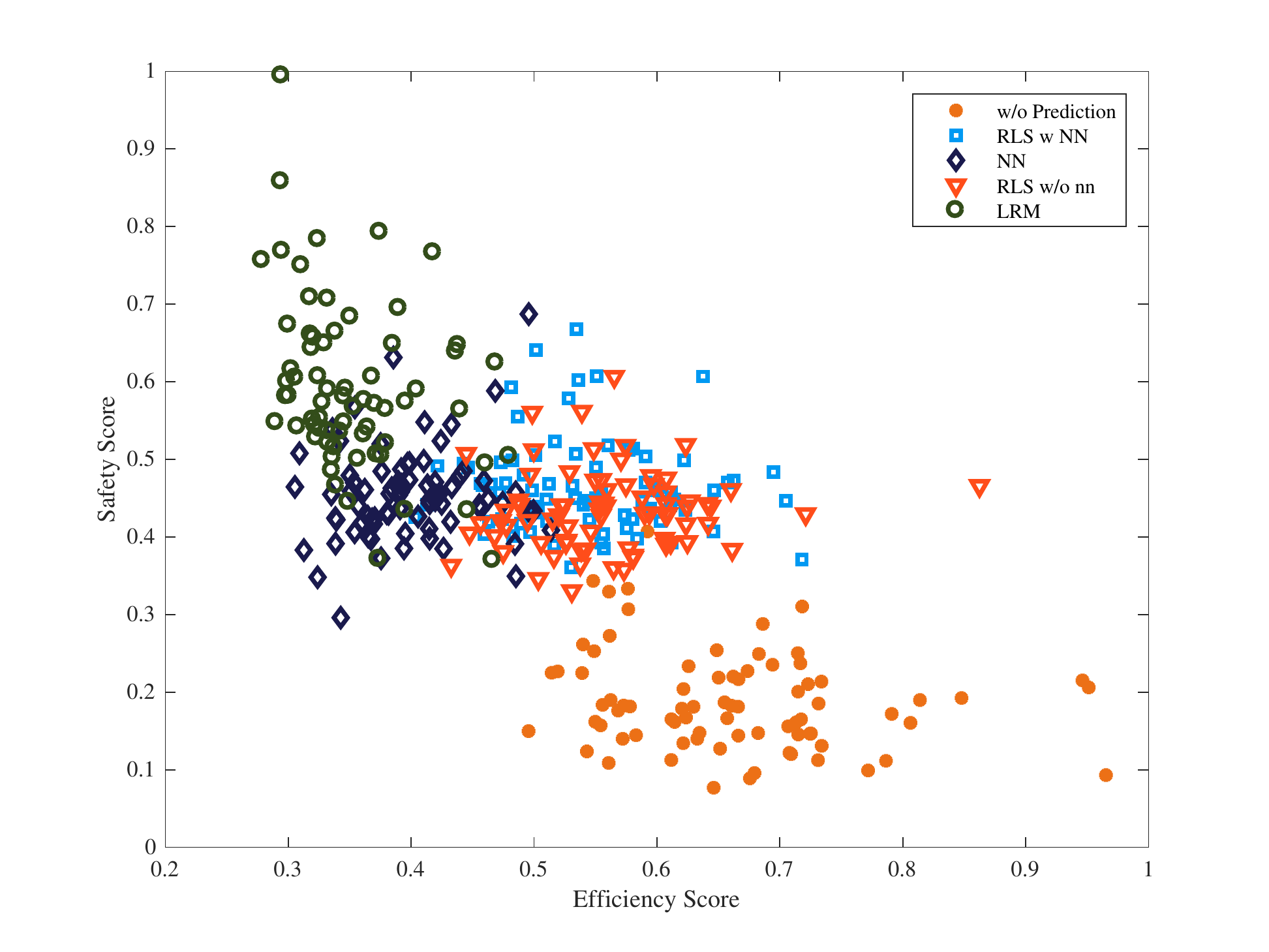}
\caption{Safety and efficiency score map for four motion prediction algorithms and collaboration system without prediction.}
\label{sne}
\end{center}
\end{figure}

We also compare the safety and efficiency scores for the four prediction methods and the scenario without prediction. Comparison results is shown in \Cref{sne}. 
Compared with the scenario without prediction, active prediction using either prediction model boosts the safety scores with a significant margin. Adaptable models achieve the same safety level as non-adaptable models, while the efficiency scores of adaptable models are not compromised. The mean and variance of safety-efficiency score of the five methods are shown in \Cref{summary_se},
Semi-adaptable neural network leads to the most robust performance with least variance, its performance is also well balanced in terms of safety and efficiency. 
\begin{table}
\caption{Mean and variance of the safety and efficiency scores for the four motion prediction models.}
 \centering
   \begin{tabular}{p{2cm}p{0.8cm}p{1cm}p{1cm}p{1cm}}
   \hline
   & safety mean & safety variance & efficiency mean & efficiency variance\\
   \hline
   RLS w/o NN & 0.5577 & 0.2593 & 0.5741 & 0.0162 \\

   RLS w NN & \bf{0.5603} & \bf{0.1353} & \bf{0.5466} & \bf{0.0055}\\

   NN  & 0.5486& 0.2775 & 0.3983 & 0.0026 \\

   LTM & 0.9209& 1.130 & 0.3541 & 0.0024\\

   w/o Prediction & 0.2550& 0.1898 & 0.6588 & 0.0095\\
\hline
   \end{tabular}

 \label{summary_se}
\end{table}

\section{Discussion}
\subsection{Hypothesis 1 - Online Adaptation Improves Accuracy}

As shown in the results, the prediction error is reduced significantly for adaptable models, which supports that online adaptable of the prediction model can improve prediction performance (\textbf{Hypothesis 1}). The quantitative metrics of prediction error mean and variance indicate semi-adaptable neural network leads to the best performance (\textbf{Hypothesis 1}). Due to high non-linearity of human motion, models parameterized by NN performances  better than models parameterized by LRM. 

Encoding nonlinear features will improve the prediction performance. Linear regression model with adaptation model only takes small quantities of past joint positions as input, and fails to encode the nonlinear features from input. Thus, small input noise can cause large prediction variance. However, semi-adaptable neural network encodes enough nonlinear features for adaptation from the raw input. Large quantities of nonlinear feature make online adaptation less sensitive to noisy input. Thus we observe small prediction error variance for semi-adaptable neural network.

\subsection{Hypothesis 2 - Effective Prediction Improves Safety}
It is notable that the average human robot distance is only \SI{0.255}{\meter} without prediction, which is less than the minimum threshold \SI{0.3}{\meter} for safe human robot interaction. However, the average safety scores are doubled when prediction modules are applied (\textbf{Hypothesis 2}). It is very common that human might approach robot at a high speed, which leaves the robot less time to escape the potential collision, especially when robot system update frequency is low. 

The prediction models predict the tendency of human motion, such that the robot can generate trajectories by taking the future constraints into consideration. When the human is moving fast toward robot, the potential collision will be detected by comparing the predicted human position and future robot trajectory. Thus replanner can be trigger in advance, and safe human robot distance is nicely kept. 

\subsection{Hypothesis 3 - Effective Prediction Improves Efficiency}
Though safety score is largely enhanced with active prediction, efficiency scores of prediction models without online adaptation are greatly compromised, since the robot's behavior is too conservative. However, when online adaptation is applied, the safety score is well maintained and the efficiency score is greatly boosted to the same level as the scenario without prediction (\textbf{Hypothesis 3}). 

Good human robot collaboration system should excel both in safety and efficiency.  When human motion prediction is good, the robot will accurately capture the tendency when human is getting away. Thus, the robot can quickly resume fetching following a new path that bypasses the predicted human trajectory. In such scenario, human safety is guaranteed and the distance between the robot and the target is also kept as close as possible. However, if human motion prediction is inaccurate, unrealistic path planning will be produced. As a result, efficiency is deteriorated.

\section{Conclusion}
In this paper, we quantitatively evaluated the effects of human motion prediction on human robot collaboration. We designed a series of human robot interaction experiments. We compared the models with different parameterizations , we also compared models with or without online parameter adaptation. The experiment results demonstrated that human motion prediction significantly enhanced the collaboration safety, and more accurate prediction led to better efficiency. Both complex parameterizations and online adaptation enhanced motion prediction performance. Adaptable prediction models that were parameterized by neural networks achieved the best and robustest performance.

\addtolength{\textheight}{-12cm}   




\bibliography{root}
\bibliographystyle{IEEEtran}

\end{document}